\documentclass[runningheads]{llncs}

\usepackage[camera-ready,year=2024,ID=10868]{eccv}

\usepackage{eccvabbrv}

\usepackage{graphicx}
\usepackage{booktabs}
\usepackage{newfloat}
\usepackage{listings}
\usepackage{bbm}
\usepackage{gensymb}
\usepackage{algorithm}
\usepackage{algorithmic}
\usepackage{multirow}
\usepackage{bbding}
\usepackage{amsmath}

\usepackage{amsthm}
\usepackage[utf8]{inputenc}
\usepackage{colortbl}
\usepackage{setspace}
\usepackage{marvosym}

\usepackage[marginal]{footmisc}

\usepackage[pagebackref,breaklinks,colorlinks,citecolor=eccvblue]{hyperref}

\begin{document}

\title{LG-Gaze: Learning Geometry-aware Continuous Prompts for Language-Guided Gaze Estimation}

\titlerunning{LG-Gaze: Language-Guided Gaze Estimation}

\author{Pengwei Yin\inst{*} \and
Jingjing Wang\inst{*}  \and
Guanzhong Zeng\inst{} \and
Di Xie\textsuperscript{\Letter}\and
Jiang Zhu\textsuperscript{\Letter}
}

\authorrunning{Y. Pengwei et al.}

\institute{Hikvision Research Institute, Hangzhou, China \\
\email{\{yinpengwei,wangjingjing9, zengguanzhong, xiedi, zhujiang.hri\}@hikvision.com}}

\maketitle

\footnote{*These authors contributed equally.\\}

\begin{abstract}
 The ability of gaze estimation models to generalize is often significantly hindered by various factors unrelated to gaze, especially when the training dataset is limited. Current strategies aim to address this challenge through different domain generalization techniques, yet they have had limited success due to the risk of overfitting when solely relying on value labels for regression. Recent progress in pre-trained vision-language models has motivated us to capitalize on the abundant semantic information available. We propose a novel approach in this paper, reframing the gaze estimation task as a vision-language alignment issue. Our proposed framework, named Language-Guided Gaze Estimation (LG-Gaze), learns continuous and geometry-sensitive features for gaze estimation benefit from the rich prior knowledges of vision-language models. Specifically, LG-Gaze aligns gaze features with continuous linguistic features through our proposed multimodal contrastive regression loss, which customizes adaptive weights for different negative samples. Furthermore, to better adapt to the labels for gaze estimation task, we propose a geometry-aware interpolation method to obtain more precise gaze embeddings. Through extensive experiments, we validate the efficacy of our framework in four different cross-domain evaluation tasks.

\end{abstract}

\section{Introduction}
\label{sec:intro}

The gaze estimation is crucial for understanding human behavior. Precise gaze estimation offers significant assistance for many applications, such as human-computer interaction \cite{8542583}, augmented reality \cite{10.1145/3084363.3085029}, and driver monitoring systems \cite{8326022}. The deep learning has led to significant improvements in gaze estimation based on appearance. While these methods show impressive results in evaluations within the same domain, they tend to experience a decline in performance when evaluated on different domains. This decline is primarily attributed to overfitting rather than learning robust features from the original domain.

\begin{figure}[ht]
	\centering
	\includegraphics[width=1\textwidth]{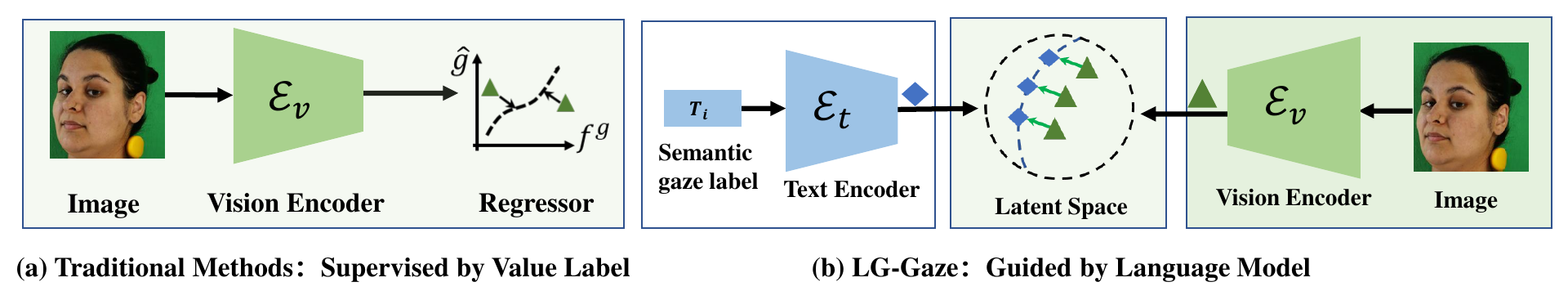}
	\caption{
		(a) The traditional method of gaze generalization involves overseeing model training by means of numerical label regression. (b) In our study, we introduce a text models to steer the development of robust features in visual models.
	}
	\label{fig:intro_cmp}
\end{figure}

Gaze data encompass diverse elements, such as appearance, wearables, environments, and image clarity\cite{gazeconsistence, Yin_Zeng_Wang_Xie_2024}, yet gaze annotations predominantly focus on eye direction, marginalizing other variables as noise. Such irrelevant factors exacerbate the domain disparity, hindering gaze models' adaptability. Additionally, supervisory numeric labels might induce overfitting, influenced by these extraneous gaze components.

As shown in Figure \ref{fig:intro_cmp}(a), training a gaze model typically involves utilizing a vision encoder to extract features and then employing a regressor for prediction. Traditional methods train a gaze model by using data disturbing \cite{gazeconsistence} ,feature enhancement \cite{cdg}, and adversarial learning \cite{cheng2022puregaze} to improve the generalization ability of the gaze model. However, these regularization techniques have limited effectiveness as they rely solely on numerical labels for supervision, making them susceptible to overfitting due to irrelevant factors. The above may indicate that  it is difficult to guide the model to learn ideal robust features through these indirect regularization terms.

To counter overfitting in gaze representation learning, we advocate for multimodal integration. Language, rich in semantics and knowledge, complements visual data. Each gaze label is described textually, \texttt{"the yaw/pitch degree angle of the person is \{yaw/pitch\}."}.
Leveraging visual-language models, such as CLIP, which excel in generalized latent space learning from vast image-text pairs, we align image features with a language space. This distillation of CLIP's language expertise acts as regularization, enhancing generalization.
However, these methods are only applicable to a limited set of discrete scalars and are not suitable for continuous tasks such as gaze estimation. Aligning gaze features to discrete text features can result in inaccurate feature alignment, which in turn affects the performance of continuous regression tasks.

To address these issues, we propose a novel training method named LG-Gaze, which is designed to learn continuous and geometry-sensitive features for gaze estimation, as depicted in Figure \ref{fig:intro_cmp}(b).
This framework aligns gaze features with continuous linguistic features extracted by a powerful language model, which not only prevents model overfitting but also enhances the  generalization capabilities of gaze model.
Primarily, unlike common discrete multimodal contrastive learning methods \cite{radford2021learning, li2022ordinalclip}, we introduce a loss function for continuous multimodal contrastive learning. Our Multimodal Contrastive Regression loss function (MCR) customizes adaptive weights for different negative samples according to label distance, facilitating more refined feature alignment in the feature space and benefiting the model in learning more continuous features.
Next, to better adapt to the vectorial property of labels for gaze estimation task, we propose a geometry-aware interpolation method to obtain more precise gaze embeddings. The geometry-aware interpolation method combines spherical \cite{spherical_inter} and bilinear interpolation techniques to resolve the issue of creating semantic labels for gaze, ensuring accurate gaze prompts.
Furthermore, MCR also resolves the limitations of previous contrastive learning functions by utilizing a more uniform distribution of global negative samples for contrast. Unlike most methods that are limited to intra-batch comparisons, our loss function benefits from extensive global contrast, resulting in a reasonable and robust feature distribution.

The main contributions can be summarized as follows:
\begin{itemize}
	\item In this paper, we propose a novel training framework called LG-Gaze for gaze estimation. LG-Gaze guides the gaze model to learn generalized representations by leveraging textual features extracted from language models, resulting in more robust gaze representations. By incorporating rich language knowledge, this approach achieves robust gaze features, ultimately enhancing the cross-domain generalization capability of the gaze model.	
    \item To learn continuous feature for regression tasks, we propose a new multimodal contrastive learning loss function named MCR. The method employ adaptive weights for different negative sample, helping learn more refined features.
    Moreover, MCR leverages ample global negative samples for contrast, leading to a reasonable feature distribution.	
    \item We introduce a geometry-aware interpolation method that employs spherical interpolation techniques to compute accurate gaze embeddings, thereby ensuring the accuracy of semantic labels.
    \item Experimental results and visualizations demonstrate that LG-Gaze not only achieves remarkable performance compared to the baseline but also surpasses state-of-the-art domain generalization methods for gaze estimation.
\end{itemize}

\section{Related Works}

\subsection{Gaze Estimation}
Appearance-based gaze estimation has garnered attention in recent years \cite{Zhang2015AppearancebasedGE,Krafka2016EyeTF,Cheng_2018_ECCV,Cheng2020GazeEB}. However, it faces challenges in cross-domain evaluation due to the domain gap resulting from various gaze-irrelevant factors. Common approaches often rely on diverse gaze datasets \cite{zhang2020eth,kellnhofer2019gaze360} for training, aiming to equip models with robust generalization capabilities. Unfortunately, collecting gaze data remains costly, and the diversity of available data remains limited. To address this issue, we propose enhancing gaze estimation models through domain generalization (DG) methods. These methods enable models to generalize to unseen distributions and enhance cross-domain performance. Most existing DG techniques are tailored for classification tasks, leaving a gap in the context of gaze estimation.

PureGaze \cite{cheng2022puregaze} debuts a self-adversarial schema, discarding irrelevant gaze cues, enhancing pertinent ones. Xu \textit{et al.} \cite{gazeconsistence} and NeRF-Gaze \cite{yin2022nerf} similarly counteract gaze distractions via adversarial data modifications and augmentation. However, residual gaze-confounding factors persist, hindering estimation precision.
CLIP-Gaze \cite{Yin_Zeng_Wang_Xie_2024} is designed to segregate gaze features from predefined textual gaze-irrelevant features, thereby enhancing the generalization capability and robustness of gaze features.
Our LG-Gaze, by directly aligning with language features, also exhibits resilience against gaze-irrelevant factors. More critically, the text features in LG-Gaze possess favorable rank properties, attributed to the geometry-aware continuous prompts learning.


\subsection{Vision Language Model}
In recent research, several studies have utilized the text manipulation and visual alignment capabilities of CLIP (Contrastive Language–Image Pretraining) \cite{radford2021learning} to improve opendetection and generalization performance in specific tasks. Notable examples include DetCLIP \cite{NEURIPS2022_3ba96055}, CLIP-Gap \cite{vidit2023clip}, and CLIP-Cluster \cite{shen2023clip}.
To further enhance the performance of vision-language models on downstream tasks, an effective strategy involves learning more contextually appropriate prompts through text prompt tuning \cite{zhou2022learning,zhou2022conditional}.
Additionally, DenseCLIP \cite{rao2022denseclip} explores the application of semantic knowledge into monocular depth estimation tasks. By matching visual features with textual features, this method achieves zero-shot monocular depth estimation.
"Teaching CLIP to Count to Ten" and CrowdCLIP propose visual language models for understanding quantities, enabling object counting and crowd estimation. While these methods enhance models' understanding of quantities, they do not address regression tasks.
OrdinalCLIP \cite{li2022ordinalclip} and L2RCLIP \cite{wang2024learning} are capable of addressing ordinal regression tasks. However, they are not as effective for continuous regression tasks.
In this paper, our method aligns gaze features with continuous linguistic features through our proposed multimodal contrastive regression loss (MCR) and the geometry-aware interpolation method.

\subsection{Contrastive Learning}
In computer vision, SimCLR \cite{simclr} and CLIP \cite{radford2021learning} pioneer contrastive loss, excelling in image classification and retrieval. Yet, these techniques aren't suited for regression, where prediction targets are continuous, not categorical. Rank-N-Contrast (RNC) \cite{zha2023rank} employs sample rankings to enhance continuity understanding. Nonetheless, RNC presumes infinite distance between negative and positive pairs, neglecting the correlation between label and feature proximities.
Contrastive Domain Generalization (CDG) \cite{cdg} leverages contrastive loss to encourage the clustering of features corresponding to similar gaze directions while separating those associated with significantly different gaze directions. However, CDG relies on empirical thresholding to define positive and negative sample pairs, which limits its applicability for continuous regression tasks. In our research, we introduce a novel contrastive loss function specifically designed for continuous regression tasks. Our approach prioritizes sample ordering while considering label distances. Unlike previous contrastive learning methods that are confined to constructing sample pairs within batches, our method demonstrates a more targeted enhancement. To elaborate further, we draw inspiration from the Momentum Contrast (MoCo) framework \cite{he2020momentum} , which utilizes a queue to store negative sample features. However, these negative samples are obtained from n steps back in time and exhibit inherent uncontrollable randomness.
Unlike delayed updates, our approach synchronously updates these samples, ensuring a controlled distribution instead of relying on randomness.

\section{Method}
\subsection{Problem Statement}
In the context of gaze estimation tasks, which fundamentally belong to regression tasks, we define the data for the source domain as $\mathcal{D}_s = {(\boldsymbol{I}_i, \boldsymbol{g}_i)}^M_{i=1}$, where $(\boldsymbol{I}_i, \boldsymbol{g}_i)$ represents the $i$-th pair of image $\boldsymbol{I}_i \in \mathbb{R}^{224\times224\times3}$ and their corresponding gaze direction $\boldsymbol{g}_i \in \mathbb{R}^{3}$. Here, $M$ denotes the total number of pairs in $\mathcal{D}_s$. Our objective in gaze estimation is to train a neural network composed of an image encoder $\mathcal{E}_v(\cdot): \mathbb{R}^{224\times224\times3} \rightarrow \mathbb{R}^d$ to extract gaze representations $\boldsymbol{f}^g_i$ based on input image $\mathcal{I}_i$ and an MLP regressor $\mathcal{R}_g(\cdot): \mathbb{R}^d \rightarrow \mathbb{R}^3$ to predict gaze directions $\boldsymbol{g}_i$. Typically, once we have trained a robust gaze representation, training $\mathcal{R}_g$ becomes relatively straightforward using regression functions (e.g., $L_1$ or AngleLoss).

However, gaze estimation research has encountered a persistent challenge that hinders its advancement: \textbf{\textit{How can we design robust gaze representations $\{\boldsymbol{\boldsymbol{f}^g_i}\}^M_{i=1}$ that are continuous, semantic, and sensitive to geometric factors for gaze estimation?}}

Although many previous methods explore additional technologies to enhance feature representations, such as adversarial training \cite{cheng2022puregaze}, contrastive learning as an auxiliary task \cite{cdg} and mitigating the impact of gaze-irrelevant factors \cite{gazeconsistence}, they often focus solely on finding regularization terms to improve model performance without explicitly aiming for robust and well-distributed features. Consequently, existing methods often face overfitting issues and struggle to achieve the desired performance. In this paper, we propose a novel language-guided gaze representation learning framework that leverages rich prior knowledge from a natural language model. Our approach aims to learn robust and well-distributed gaze representations, addressing the aforementioned challenges.

\begin{figure}[ht]
	\centering
	\includegraphics[width=1\textwidth]{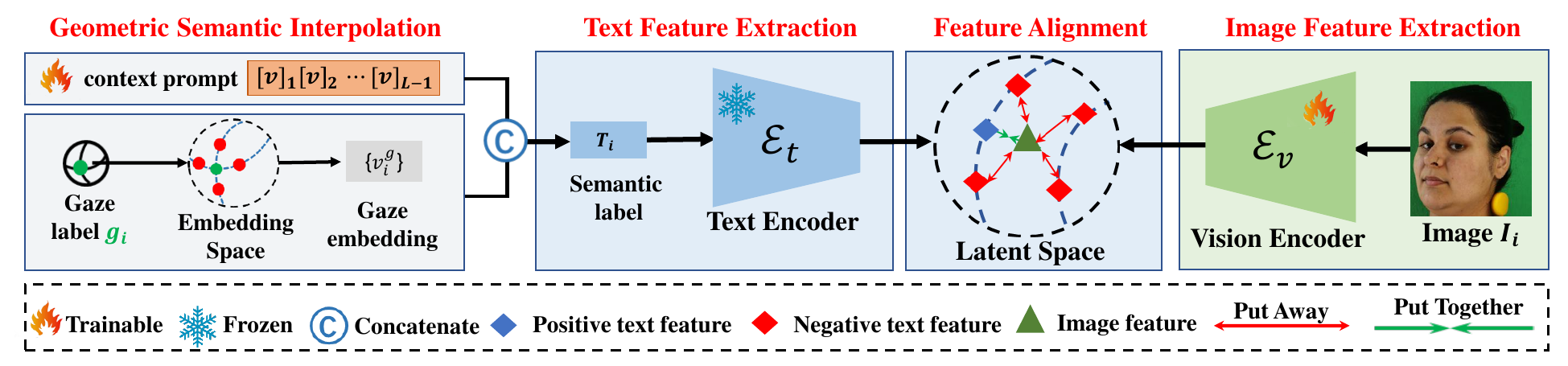}
	\caption{We reformulate the task as an image-language matching problem, which mainly consists of a trainable prompt, a frozen text encoder, a trainable image encoder.}
	\label{fig:framework}
\end{figure}

\subsection{LG-Gaze Framework}
To harness the full power of language, we utilize the text encoder from CLIP as our language model. CLIP adopts a vision-language pre-training framework and learns representations from image-text pairs, thereby constructing a joint vision-language latent space. Given the strong performance of the CLIP model in downstream tasks, we implemented our method using the original CLIP text encoder from the referenced paper.

Suppose that each training step is provided with a data batch ${(\boldsymbol{I}_i, \boldsymbol{g}_i)}^B_{i=1}$, where $\boldsymbol{g}_i$ represents the gaze label of the image $\boldsymbol{I}_i$. Our goal is to construct semantic labels and use a language model to extract text features. This process will help the gaze model  learn robust features by aligning them with the extracted features. To leverage the prior knowledge within text features, we begin by constructing a learnable prompt $\{\boldsymbol{v}_1,\boldsymbol{v}_2, \dots,\boldsymbol{v}_{L-1}\} \in \mathbb{R}^{512\times (L-1)}$ as the context for describing gaze estimation tasks instead of handcrafting the prompt context. Simultaneously, we treat gaze labels as individual words. For each sample, we concatenate the learnable prompt with the embedding $\boldsymbol{\boldsymbol{v}_i^g} \in \mathbb{R}^{512}$ to form a sequence $\boldsymbol{T_i} = \{\boldsymbol{v}_1,\boldsymbol{v}_2, \dots,\boldsymbol{v}_{L-1}, \boldsymbol{\boldsymbol{v}_i^g}\}$ as a semantic label. Specifically, we utilize the gaze embedding as the final token in the sequence $\boldsymbol{T_i} \in \mathbb{R}^{512 \times L}$.

To ensure the continuity, ordinality, and geometric properties of gaze label embeddings, we introduce $N$ learnable embedding anchors, denoted as $\{\boldsymbol{A_j}\}^N_{j=1}$. Each anchor corresponds to a unique gaze vector, uniformly distributed across the entire label space. We utilize a nearest-neighbor interpolation method to represent gaze label embeddings associated with sample relationships. This approach yields a continuous and geometry-aware representation of gaze embeddings, as detailed in Section \ref{sec:semantic_label}.

In our research, we send image $\boldsymbol{I}_i$ to an image encoder $\mathcal{E}_v$ to extract the image feature $\boldsymbol{f}^g_i$. Consequently, the features of a batch can be represented as $\{\boldsymbol{f}^g_i\}^B_{i=1}$. Simultaneously, we utilize the language model $\mathcal{E}_t$ to obtain corresponding text features $\{\boldsymbol{f}^t_i\}^B_{i=1}$ from $\{\boldsymbol{T_i}\}^B_{i=1}$. While the parameters of the language model remain frozen, we train the entire image encoder to align image features with the language feature space. Subsequently, we utilize an image-text contrastive loss to optimize the network, which includes both an image-to-text loss and a text-to-image loss, following the methodology of CLIP. Detailed information about our proposed contrastive learning method will be provided in Section \ref{sec:fea_align}.

After aligning the features, we obtain robust gaze representations.
Finally, the image features are fed into a regressor denoted as $\mathcal{R}_g$, yielding the predicted gaze direction $\hat{g}_i \in \mathbb{R}^3$. Specifically, we predict  through supervised learning, which is defined as:

\begin{equation}
	\label{loss:l_gaze}
	\mathcal{L}_{Gaze}\left(\hat{\boldsymbol{g}_{i}},\boldsymbol{g}_{i}\right)
	=
	\arccos
		\left(
		\frac
			{
				\hat{\boldsymbol{g}_{i}}
				\cdot
				\boldsymbol{g}_{i}
			}
			{
				\|\hat{\boldsymbol{g}_{i}}\|
				\|\boldsymbol{g}_{i}\|
			}
		\right)
\end{equation}

\begin{figure}[ht]
	\centering
	\includegraphics[width=0.9\textwidth]{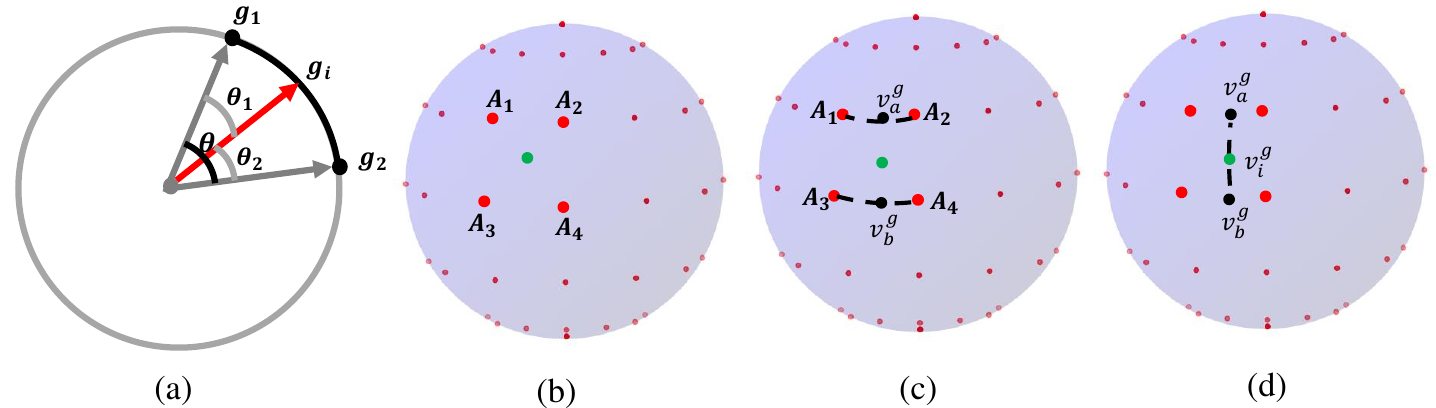}
	\caption{These four subfigures contain all steps of our proposed interpolation method.}
	\label{fig:interpolation}
\end{figure}

\subsection{Geometry-aware Continuous Gaze Prompts}
\label{sec:semantic_label}

In this section, to better represent gaze labels in the embedding space, we construct a gaze embedding for each label. This embedding should have the capability for continuous representation and geometric attributes. To address the conflict between the finite vocabulary in natural language processing and the infinite nature of gaze labels, we interpolate each gaze embedding $\boldsymbol{v}^g_i$ through $N$ gaze anchor embeddings $\{\boldsymbol{A_j}\}^N_{j=1}$.

Common approaches, such as those described in \cite{li2022ordinalclip}, employ global linear interpolation to represent $\boldsymbol{v}^g_i$, which interpolates the target embedding using all anchor embeddings. The weight $w_{i,j}$ for each anchor embedding is computed by Equation \ref{eq:global_eq}, determined by the similarity between its corresponding gaze vector $\boldsymbol{g}_j$ and the target $\boldsymbol{g}_i$. However, global interpolation cannot accurately represent $\boldsymbol{v}^g_i$ because anchors that are not proximate fail to provide effective information.

\begin{equation}
\label{eq:global_eq}
w_{i,j}=\frac{\cos \left(\boldsymbol{g}_i, \boldsymbol{g}_j\right)}{\sum_{j=1}^N \cos \left(\boldsymbol{g}_i, \boldsymbol{g}_j\right)}
\end{equation}

To further enhance interpolation, we attempted to employ a bilinear interpolation method. However, as it still utilizes a linear representation, this method remains insufficiently precise in expressing the geometric relationships between gaze vectors.

To address this issue, we decided to adopt a spherical interpolation method \cite{spherical_inter}, which can accurately calculate the interpolation weights between vectors. As illustrated in Figure \ref{fig:interpolation}(a), a flat spherical interpolation method is depicted. For the vector $\boldsymbol{g}_i$ positioned between $\boldsymbol{g}_1$ and $\boldsymbol{g}_2$, we initially compute $\theta = \arccos(\langle \boldsymbol{g}_1, \boldsymbol{g}_2 \rangle)$ and $\theta_1 = \arccos(\langle \boldsymbol{g}_1, \boldsymbol{g}_i \rangle)$. Subsequently, we compute the scalar $t$ by $t={\theta_1} / \theta$ ,then we apply the spherical interpolation Equation \ref{eq:spherical_2d} to determine the corresponding weights for $\boldsymbol{g}_1$ and $\boldsymbol{g}_2$. Finally, we can represent the target vector $\boldsymbol{g}_i$ through the summation of these weights. Ultimately, we confirm that this method of weight calculation is accurate.

\begin{equation}
	\begin{gathered}
	\label{eq:spherical_2d}
		\boldsymbol{g}_i=\frac{\sin ((1-t) \theta)}{\sin (\theta)} \boldsymbol{g}_1+\frac{\sin (t \theta)}{\sin (\theta)} \boldsymbol{g}_2
	\end{gathered}
\end{equation}

Building upon the aforementioned spherical interpolation method, we further integrate the concept of bilinear interpolation to derive the interpolation weights for each gaze label on a stereographic sphere. As depicted in Figure \ref{fig:interpolation}(b), we initially identify the four nearest anchor embeddings $\boldsymbol{A}_{1-4}$ based on their proximity. Subsequently, as shown in Figure \ref{fig:interpolation}(c), we calculate the weights $w_{a,1}$ and $w_{a,2}$ for $\boldsymbol{A}_1$ and $\boldsymbol{A}_2$ using Equation \ref{eq:spherical_2d}. Similarly, we can determine $w_{b,3}$ and $w_{b,4}$. In Figure \ref{fig:interpolation}(d), we proceed to obtain $w_{i,a}$ and $w_{i,b}$ using a similar method. Ultimately, we acquire the associated weights for $\boldsymbol{A}_{1-4}$ to interpolate $\boldsymbol{v}^g_i$, which can be expressed as follows:
\begin{equation}
\boldsymbol{v}^g_i=(w_{i, a} * w_{a, 1})\cdot \boldsymbol{A}_1 + \
	  (w_{i, a} * w_{a, 2})\cdot \boldsymbol{A}_2 + \
	  (w_{i, b} * w_{b, 3})\cdot \boldsymbol{A}_3 + \
	  (w_{i, b} * w_{b, 4})\cdot \boldsymbol{A}_4
\end{equation}

Additionally, to enhance the geometric relationships between gaze embeddings, we employ a simple loss function to constrain anchor embeddings $\{\boldsymbol{A_j}\}^N_{j=1}$. As shown in Equation \ref{eq:geo}, the similarity between anchor embeddings should be consistent with their relationships in the label space.
\begin{equation}
\label{eq:geo}
\ell_{\text {Geo}}=\frac{1}{M*M} \sum_{i=1}^M \sum_{j=1}^M L_1\left(\cos \left(\boldsymbol{A}_i, \boldsymbol{A}_j\right), \cos \left(\boldsymbol{g}_i, \boldsymbol{g}_j\right)\right)
\end{equation}

\subsection{Continuous Multimodal Contrastive Regression Loss Function}
\label{sec:fea_align}

Mathematically, from the aforementioned steps, we obtain image features $\{\boldsymbol{f}^g_i\}^B_{i=1}$ and their corresponding text features $\{\boldsymbol{f}^t_i\}^B_{i=1}$. Both CLIP and OrdinalCLIP utilize the InfoNCE loss to align the multimodal features. To delve into specifics, consider an image-text feature pair $(\boldsymbol{f}^g_i, \boldsymbol{f}^t_i)$. Here, $\boldsymbol{f}^g_i/\boldsymbol{f}^t_i$ serves as the positive sample for $\boldsymbol{f}^t_i/\boldsymbol{f}^g_i$, as they are associated with the same gaze label. Meanwhile, all other image/text features in a mini-batch are treated as negative samples and are consequently pushed away from $\boldsymbol{f}^t_i/\boldsymbol{f}^g_i$. However, this training objective poses challenges for regression tasks because it overlooks the finer-grained distance ordering relationships between samples. Consequently, failing to differentiate the relative distances between image/text and their corresponding negative samples will inevitably weaken the learning effectiveness, especially in cross-modal representation learning for regression tasks.

Considering the proximity relationships between data, for the $i$-th sample, we introduce a weight parameter $\{w(i,j)\}^B_{j=1}$ for each negative sample. This parameter is computed based on the  distances between labels of samples. Our proposed multimodal contrastive regression loss (MCR) effectively controls the feature distances among samples, preventing negative samples from being excessively contracted. As a result, we learn a semantically meaningful representation space tailored for regression tasks, ensuring consistency between sample feature relationships and the label space. The proposed text-to-image MCR loss is formulated as follows.
\begin{equation}
\label{loss:t2i}
 \ell_{t2i} =
 -\frac{1}{B}\sum_{i=1}^B \log
 \left(
	 \frac
		 {
		 	\exp \left(\operatorname{sim}\left(\boldsymbol{f}_i^t, \boldsymbol{f}_i^g\right)\right)
		 }
		 {
		 	\exp
		 	\left(
		 		\operatorname{sim}
		 			\left(
		 				\boldsymbol{f}_i^t, \boldsymbol{f}_i^g
		 			\right)
		 	\right)
		 	+
		 	\sum_{j=1}^B w(i, j) \cdot \exp
		 		\left(
		 			\operatorname{sim}
		 				\left(
		 					\boldsymbol{f}_i^t, \boldsymbol{f}_j^g
		 				\right)
		 		\right)
		 }
 \right)
\end{equation}
where ${sim}\left(\boldsymbol{f}_i^t, \boldsymbol{f}_i^g\right) = cos(\boldsymbol{f}_i^t, \boldsymbol{f}_i^g)$ and $w(i,j)=cos(\boldsymbol{g}_i, \boldsymbol{g}_j)$ denotes the contrastive weight of the $j$-th negative image sample with respect to the $i$-th text feature in our LG-Gaze framework. The weight parameter should be directly proportional to the label distance between negative samples. As the  distance between labels increases, the penalty on negative sample weights should also increase. This encourages feature embeddings to adhere to real-world relationships.

The MCR loss for image-to-text can be formulated in a similar manner. However, what sets it apart is our deliberate expansion of the number of negative samples to facilitate the learning of a robust representation. To achieve this, we draw inspiration from MoCo, which utilizes a dynamic queue to maintain a diverse set of negative samples, thereby enhancing the effectiveness of contrastive learning. Nevertheless, this approach still faces some challenges. For instance, if the queue update rate is too slow, the negative samples stored in the queue may become outdated and fail to align with the current model. Additionally, due to the inherent randomness and limited diversity of negative samples, there could be discrepancies between the queue data and the entire data distribution. To address this issue, we utilize the flexibility of language labels by creating $K$ global text features $\{\boldsymbol{\boldsymbol{f}^t_i}\}^K_{i=1}$ as negative samples (where $K$ is significantly larger than $B$). These text labels correspond to gaze vectors that show a uniformly dense distribution in the label space. Furthermore, these $K$ text labels are interpolated based on $N$ anchor $\{\boldsymbol{\boldsymbol{A}_j}\}^N_{j=1}$. As a result, the proposed image-to-text MCR loss can be formulated as follows:
\begin{equation}
\label{loss:i2t}
\ell_{i2t} =
-\frac{1}{B} \sum_{i=1}^B \log
\left(
	\frac
	{
		\exp \left(\operatorname{sim}\left(\boldsymbol{f}_i^g, \boldsymbol{f}_i^t\right)\right)
	}
	{
		\exp
			\left(
				\operatorname{sim}
					\left(
						\boldsymbol{f}_i^g, \boldsymbol{f}_i^t
					\right)
			\right)
		+
		\sum_{j=1}^{B+K} w(i, j)
		\cdot
		\exp
			\left(
				\operatorname{sim}
					\left(
						\boldsymbol{f}_i^g, \boldsymbol{f}_j^t
					\right)
			\right)
	}
\right)
\end{equation}

Obviously, we can observe that the difference between $\ell_{i2t}$ and $\ell_{t2i}$ lies in the number of negative samples. We concatenate $K$ global negative samples with the $B$ negative samples from mini-batches to form the complete set of negative samples. Through this operation, the global negative samples remain relatively fixed, mitigating issues related to randomness. Additionally, their uniform distribution helps alleviate discrepancies between the negative samples and the entire data distribution. Furthermore, by computing features based on the current model, we address potential outdated information. Finally, our total MCR loss can be denoted as:
\begin{equation}
\label{loss:multi_modal_cl}
	\ell_{MCR}=\ell_{i 2 t}+\ell_{t 2 i}
\end{equation}

\subsection{Overall Objective}
In summary, the overall objective implemented in our framework is:
\begin{equation} \mathcal{L} = \lambda_{1} \ell_{Geo}  + \lambda_{2}\ell_{MCR}  + \lambda_{3}\ell_{Gaze}
\nonumber
\end{equation}

$\lambda_{1}, \lambda_{2}, \lambda_{3}$ are hyperparameters, and we empirically set $\lambda_{1} = \lambda_{2} = \lambda_{3} = 1.0$.

\section{Experiments}

\subsection{Experiment Details }

\subsubsection{Data Preparation}
We evaluate gaze estimation method on four cross-domain tasks, utilizing ETH-XGaze \cite{zhang2020eth} and Gaze360 \cite{kellnhofer2019gaze360} as the training datasets, and MPIIFaceGaze \cite{zhang2017mpiigaze} and Eye-Diap \cite{funes2014eyediap} as the test datasets. We denote them as $\mathcal{D}_\mathrm{E}$ (ETH-XGaze)$\rightarrow$$\mathcal{D}_\mathrm{M}$ (MPIIFaceGaze), $\mathcal{D}_\mathrm{E}$$\rightarrow$$\mathcal{D}_\mathrm{D}$(EyeDiap),
 $\mathcal{D}_\mathrm{G}$(Gaze360)$\rightarrow$$\mathcal{D}_\mathrm{M}$,   $\mathcal{D}_\mathrm{G}$$\rightarrow$$\mathcal{D}_\mathrm{D}$. See the supplementary material for more details.

\subsubsection{Vision Model Implementation Details}
We use ResNet-18 as our gaze feature extractor $\mathcal{E}_v$, and a fully connected layer to regress a 3-dimensional gaze vector. We resize and normalize all the images to 224$\times$224 pixels and scale the pixel values between 0 and 1. We set the batch size to 64. We maintain the same setup for training the source domain models on $\mathcal{D}_E$ and $\mathcal{D}_G$.

\subsubsection{Language Model Implementation Details}
The architecture of the text encoder is adapted from the Transformer \cite{vaswani2017attention}, incorporating modifications outlined by Radford \cite{radford2019language}. The dimensions for both text and gaze features have been standardized to 1024. Our models are constructed using the foundational open-source code of CLIP. The class token is strategically placed at the terminal position of the sequence, with the context tokens quantity designated as $L=10$. Additionally, we set the yaw angle range for the anchors to be [$-180^\circ$, +179$^\circ$]  and the pitch range to be [-90$^\circ$, +90$^\circ$]  with a step size of 30$^\circ$, resulting in a total of 91 anchors. We set the number of negative samples to 256, which are uniformly distributed in the gaze label space using the Fibonacci sphere algorithm \cite{gonzalez2010measurement}.

\begin{table*}[ht]
	\renewcommand\arraystretch{1.0}
	\setlength{\tabcolsep}{2.0mm}
	\centering
	\scalebox{1.0}{
		\begin{tabular}{|l|l|c|c|c|c|c|c|c|}
			\hline
			\multirow{2}{*}{\scalebox{1.0}{Task}}& 	\multirow{2}{*}{\scalebox{1.0}{Methods}}& \multirow{2}{*}{\scalebox{1.0}{$\mid\mathcal{D}_t \mid$}}&
			\scalebox{1.0}{$\mathcal{D}_\mathrm{E}  $} &
			\scalebox{1.0}{$\mathcal{D}_\mathrm{E}  $} &
			\scalebox{1.0}{$\mathcal{D}_\mathrm{G}  $} &
			\scalebox{1.0}{$\mathcal{D}_\mathrm{G}  $} &
			\multirow{2}{*}{Avg}
			\\
			& 	 &  &
			\scalebox{1.0}{$ \rightarrow \mathcal{D}_\mathrm{M}$} &
			\scalebox{1.0}{$ \rightarrow \mathcal{D}_\mathrm{D}$} &
			\scalebox{1.0}{$ \rightarrow \mathcal{D}_\mathrm{M}$} &
			\scalebox{1.0}{$ \rightarrow \mathcal{D}_\mathrm{D}$} &
			\\
			\hline \hline
			\multirow{6}{*}{ \scalebox{1.0}{DG} }
			& \scalebox{1.0}{CNN Baseline} & 0 & 8.47 & 9.32 & 7.54 & 8.93 & 8.57 \\
			& \scalebox{1.0}{PureGaze\cite{cheng2022puregaze}} & 0 & 7.08 & 7.48 & 9.28 & 9.32 & 8.29 \\
			& \scalebox{1.0}{CDG\cite{cdg}} $^\ddagger$ & 0 & 6.73 & 7.95 & 7.03& 7.27& 7.25\\
			& \scalebox{1.0}{Xu $\textit{et al.}$}\cite{gazeconsistence} &  0 & 6.50& 7.44& 7.55 & 9.03 & 7.63 \\
                & \scalebox{1.0}{CLIP-Gaze}\cite{Yin_Zeng_Wang_Xie_2024} &  0&  \textbf{6.41}& 7.51& 6.89 & 7.06 & 6.97\\
			& \scalebox{1.0}{Our LG-Gaze} &  0 & 6.45& \textbf{7.22} & \textbf{6.83} & \textbf{6.86} & \textbf{6.84} \\
			\hline \hline
			\multirow{6}{*}{\scalebox{1.0}{UDA}}
			& \scalebox{1.0}{PnP-GA ${^*}$}\cite{pnpga} & 10 & 5.53 & 5.87 & 6.18 & 7.92 & 6.38 \\
			& \scalebox{1.0}{RUDA}\cite{ruda} & 100 & 5.70 & 6.29 & 6.20 & 5.86 & 6.01 \\
			& \scalebox{1.0}{CRGA}\cite{cdg} & $>0$ & 5.48 & 5.66 & 5.89 & 6.49 & 5.88 \\
			& \scalebox{1.0}{LatentGaze}\cite{latentgaze} & $100$ & 5.21 & 7.81 & - & - & 6.51 \\
			& \scalebox{1.0}{Liu $\textit{et al.}$}\cite{jitter} & 100 & 5.35 & 6.62 & 7.18 & 8.61 & 6.94 \\
			& \scalebox{1.0}{UnReGA}\cite{unrega} & 100 & 5.11 & 5.70 & 5.42 & 5.80 & 5.51 \\
			\hline
	\end{tabular}
 }
	
	\caption{Comparison with state-of-the-art methods. We report the results by angular error in degrees, and use bold and underline to indicate the best and the second best result in each column for a specific task. $^\ddagger$ means the model uses ResNet-50 as backbone, ${^*}$ shows that the experimental settings are different.}
	\label{tab:full_comp}
\end{table*}

\subsubsection{Training Implementation Details}
We conducted the experiments on a single Tesla V100 GPU. Specifically, we use the SGD optimizer with Nesterov momentum, a learning rate (LR) of 5$\times$10$^{-2}$ and a weight decay of 1$\times$10$^{-5}$ for the parameters. We train for 30 epochs using a Cosineannealing LR scheduler \cite{loshchilov2016sgdr} with a 3-epoch warm-up. We apply a data augmentation technique involving a random color field and grayscale, as described in \cite{mocov2}.

\subsection{Performance Comparison with SOTA Gaze Estimation Methods}
\subsubsection{Comparison of Cross-domain Evaluation.}
Table \ref{tab:full_comp} presents the quantitative results of four cross-domain tests. The second row compares our LG-Gaze with the SOTA domain DG methods for gaze estimation. The CNN baseline method refers to the one that  relies solely on the vision encoder to extract the single visual modality feature and uses $\ell_{gaze}$ for supervision. In summary, the LG-Gaze achieves the best overall performance and demonstrates state-of-the-art results on three cross-domain evaluation tasks. It also achieves the second-best performance for $\mathcal{D}_\mathrm{E}$ $\rightarrow$ $\mathcal{D}_\mathrm{M}$, highlighting the effectiveness of our method.

\begin{figure}[h!]
	\centering
	\includegraphics[width=0.8\textwidth]{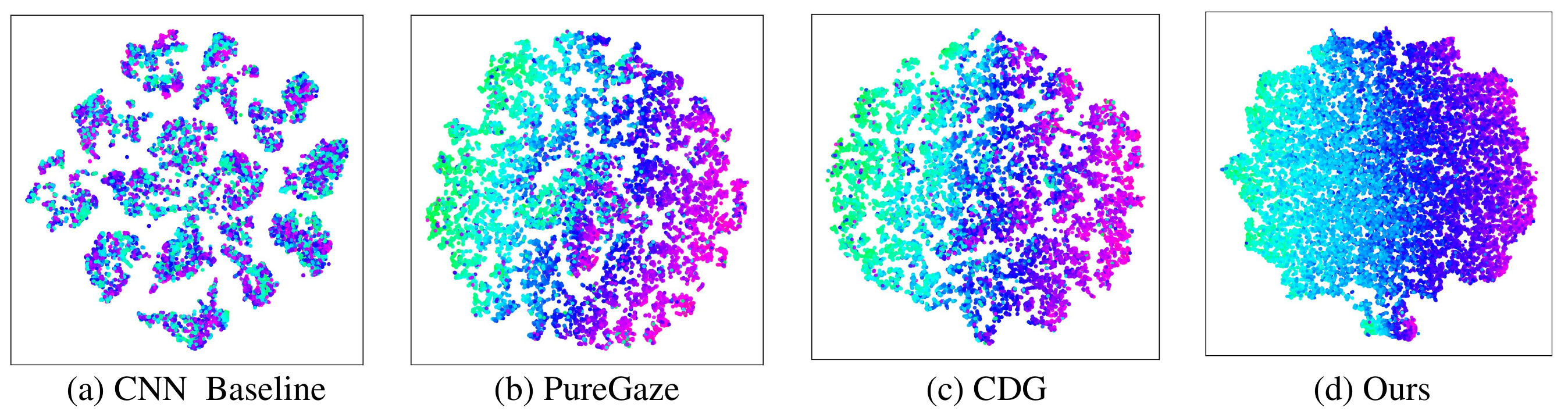}
	\caption{Illustration of the feature distribution. Various colors indicate different gaze directions and similar gaze directions have similar colors. (Recommend viewing in color).}
	\label{fig:gaze_features_vis}
\end{figure}

In addition, we present a comparison with SOTA unsupervised domain adaptation (UDA) methods in the third row of Table \ref{tab:full_comp}. Note that UDA methods require the use of a small batch of unlabeled target domain samples, which is more costly than DG methods. It can be clearly seen from the table that our LG-Gaze achieves comparable performance without extra target domain data. Especially on the two tasks of $\mathcal{D}_\mathrm{G}$ $\rightarrow$ $\mathcal{D}_\mathrm{M}$ and $\mathcal{D}_\mathrm{G}$ $\rightarrow$ $\mathcal{D}_\mathrm{D}$, it is close to UDA methods. Therefore, this verifies the powerful performance of our proposed method, as it does not rely on any prior information from the target domain.

\begin{table}[h!]
	\renewcommand\arraystretch{1.1}
	\setlength{\tabcolsep}{1.8mm}
	\centering
	\scalebox{1}{
		\begin{tabular}{|l|c|c|c|c|c|}
			\hline
			Methods&
			$\mathcal{D}_\mathrm{E} \rightarrow \mathcal{D}_\mathrm{M}$&
			$\mathcal{D}_\mathrm{E} \rightarrow \mathcal{D}_\mathrm{D}$&
			$\mathcal{D}_\mathrm{G} \rightarrow \mathcal{D}_\mathrm{M}$&
			$\mathcal{D}_\mathrm{G} \rightarrow \mathcal{D}_\mathrm{D}$&
			Avg\\
			\hline
			\hline
			CNN Baseline& $8.35$  & $\underline{9.66}$  & $\underline{7.58}$  & $\underline{9.01}$  &  $\underline{8.65}$ \\
			Vanilla CLIP \cite{radford2021learning}& $82.43$ & $79.64$ & $82.43$ & $79.64$ &  81.04  \\ 	
			CoOp \cite{zhou2022learning} &        $9.23$  & $11.06$ & $13.15$ & $10.85$ &  $11.07$\\
			OridinalCLIP \cite{li2022ordinalclip}& $\underline{7.88}$ & 10.11	 & 9.12 & 	9.74 & 	9.21 \\
			Our LG-Gaze& $\textbf{6.45}$ & $\textbf{7.22}$ & $\textbf{6.83}$ & $\textbf{6.86}$  & $\textbf{6.84}$ \\
			\hline
	\end{tabular}}
	\caption{Comparisons to SOTA contrastive learning methods. Bold indicates the best results in each column, and underline denote the second best result in each column.}
	\label{tab:VLM_comp}
\end{table}

\subsubsection{Comparison of Feature Visualization.}

To evaluate and analyze the gaze features $\boldsymbol{f}^{g}$ extracted by different models, we follow the same scheme as CDG \cite{cdg}, and visualize the gaze features of the cross-domain task $\mathcal{D}_\mathrm{G}$$\rightarrow$$\mathcal{D}_\mathrm{D}$ using t-SNE \cite{tsne} to compare different methods. Figure \ref{fig:gaze_features_vis} shows the visualization results from all DG methods in the second row of Table \ref{tab:full_comp}, where feature points with similar gaze labels have similar colors. For the baseline model, the features of different gaze labels are mixed together, which is unreasonable for the regression task as shown in Figure \ref{fig:gaze_features_vis} (a). The ideal feature distribution should be continuous and gradual with labels for the regression task \cite{zha2024rank}. Figure \ref{fig:gaze_features_vis} (b,c) represents several classic DG methods that have been enhanced compared to the baseline, but they exhibit unreasonable feature distributions. In general, LG-Gaze has the most reasonable feature distribution, and the visualization result is shown in Figure \ref{fig:gaze_features_vis} (d). The strong correlation between gaze direction and features indicates that guiding  visual feature learning through language is effective.

\subsection{Performance Comparison with SOTA VLM Regression Methods}
Table \ref{tab:VLM_comp} presents the performance of various methods that employ VLM models. To investigate the applicability of ordinal regression methods  to gaze estimation, we discretize the gaze labels into integers. This mapping allows the yaw and pitch in the gaze label to be categorized into a finite number of categories. During testing, we predict yaw and pitch separately, but the accuracy of the gaze vector, which is composed of yaw and pitch, is evaluated.

\begin{table}[ht]
	\renewcommand\arraystretch{1.1}
	\setlength{\tabcolsep}{2mm}
	\centering
	\scalebox{1}{
		\begin{tabular}{|l|c|c|c|c|c|}
			\hline
			Methods&
			$\mathcal{D}_\mathrm{E} \rightarrow \mathcal{D}_\mathrm{M}$&
			$\mathcal{D}_\mathrm{E} \rightarrow \mathcal{D}_\mathrm{D}$&
			$\mathcal{D}_\mathrm{G} \rightarrow \mathcal{D}_\mathrm{M}$&
			$\mathcal{D}_\mathrm{G} \rightarrow \mathcal{D}_\mathrm{D}$&
			Avg\\
			\hline
			\hline
			CNN Baseline & $8.35$ & $9.66$ & $7.58$ & $9.01$  & $8.65$ \\
			KL-Loss &  6.77& 	7.36&	6.97&	7.08&	7.05  \\
			CDG \cite{cdg}&  6.98& 	7.45&	7.14&	7.18&	7.19   \\
			RNC \cite{zha2023rank}&  6.63&	$\underline{7.28}$&	6.99&	$\underline{6.96}$&	6.97   \\
			MoCo \cite{he2020momentum}& $\underline{6.52}$& 	7.33&	$\underline{6.95}$&	7.04&	$\underline{6.96}$  \\
			MCR& $\textbf{6.45}$ & $\textbf{7.22}$ & $\textbf{6.83}$ & $\textbf{6.86}$  & $\textbf{6.84}$ \\
			\hline
	\end{tabular}}
	\caption{Comparison of different text prompt tuning methods for gaze model cross-domain evaluation in four tasks.
		The rest of the settings are consistent for all methods, except for the contrastive loss function.
		Bold indicates the best results in each column, and underline denote the second best result results in each column.}
	\label{tab:cr_comp}
\end{table}

For Vanilla CLIP, there is no need to use source data for model training. Instead, a zero-shot method is used for prediction, resulting in only two outcomes.
First, we construct the text prompt
\texttt{"gaze estimation: the yaw/pitch degree angle of the person is \{yaw/pitch\}."}
Then, we obtain the model's prediction of yaw and pitch by maximizing the alignment between text features and image features, which is essentially in line with the original image-text matching.
However, we can observe from Table \ref{tab:VLM_comp} that Vanilla CLIP exhibits poor performance, possibly due to its struggle in distinguishing ordinal concepts.
For CoOp, the gaze prediction performance can be greatly enhanced through text prompt tuning, but it is weaker than the CNN baseline. This could be due to its inability to learn the ordinal relationship between samples.
For OrdinalCLIP, although it can learn the ordinal relationship, the  discretization of labels may lead to errors.
Additionally, following the methodology of OrdinalCLIP, we have created a visualization of the semantic labels constructed by different VLM methods in the supplementary material.

\subsection{Performance Comparison with SOTA Contrastive Regression Loss Functions}
Table \ref{tab:cr_comp} presents various methods for feature alignment or contrastive learning in gaze estimation.
For KL-Loss, it primarily constrains the similarity matrix between gaze features and text features, as well as the similarity matrix composed of gaze labels.
We follow the ideas of CDG \cite{cdg}, RNC \cite{zha2023rank}, and MoCo \cite{he2020momentum}, and transform them into image-to-text and text-to-image contrastive loss functions similar to LG-Gaze.
From Table \ref{tab:cr_comp}, we can see that our MCR loss function outperforms other methods significantly, attributed to the inclusion of constructed global negative samples and the weight coefficient for negative samples.

\subsection{Ablation Study}

\subsubsection{Effectiveness of Our Framework.}
Table \ref{tab:loss_func} shows the performance of different combinations of loss functions.
Keeping only $\ell_{Gaze}$ is equivalent to CNN baseline, and combining all the losses is equivalent to full LG-Gaze.
Compared with CNN baseline, using MCR loss to learn robust features with language guidance can significantly improve the performance.
Further adding $\ell_{Geo}$ enhances the geometric relevance between prompts, and thus improves the model performance.

\subsubsection{Effectiveness of Interpolation Methods.}
To compare the performance of global linear interpolation, bilinear interpolation and our proposed geometry-aware spherical interpolation.
Our method demonstrates superior performance due to its consideration of geometric accuracy, as shown in Table \ref{tab:interpolation}.

\begin{table}[ht]
	\renewcommand\arraystretch{1.0}
	\setlength{\tabcolsep}{1.2mm}
	\centering
	\scalebox{1}{
		\begin{tabular}{|ccc|c|c|c|c|c|}
			\hline
			$\ell_{Geo}$& 	$\ell_{MCR}$& 	$\ell_{Gaze}$&
			$\mathcal{D}_\mathrm{E} \rightarrow \mathcal{D}_\mathrm{M}$&
			$\mathcal{D}_\mathrm{E} \rightarrow \mathcal{D}_\mathrm{D}$&
			$\mathcal{D}_\mathrm{G} \rightarrow \mathcal{D}_\mathrm{M}$&
			$\mathcal{D}_\mathrm{G} \rightarrow \mathcal{D}_\mathrm{D}$&
			Avg\\
			\hline
			\scalebox{0.7}{\XSolid}&  \scalebox{0.7}{\XSolid}& \Checkmark& 8.35& 9.66&	7.58& 9.01&	8.65 \\
			\scalebox{0.7}{\XSolid}&  \Checkmark& \Checkmark& 6.76& 	7.35&	6.93&	6.98&	7.01 \\
			\Checkmark&  \Checkmark& \Checkmark& $\textbf{6.45}$ & $\textbf{7.22}$ & $\textbf{6.83}$ & $\textbf{6.86}$  & $\textbf{6.84}$ \\
			\hline
		\end{tabular}
	}
	\caption{Ablation study on loss functions. Results are reported by in degrees.}
	\label{tab:loss_func}
\end{table}

 \begin{table}[h!]
 	\renewcommand\arraystretch{1.0}
 	\setlength{\tabcolsep}{1.8mm}
 	\centering
 	\scalebox{1}{
 		\begin{tabular}{|l|c|c|c|c|c|}
 			\hline
 			Methods&
 			$\mathcal{D}_\mathrm{E} \rightarrow \mathcal{D}_\mathrm{M}$&
 			$\mathcal{D}_\mathrm{E} \rightarrow \mathcal{D}_\mathrm{D}$&
 			$\mathcal{D}_\mathrm{G} \rightarrow \mathcal{D}_\mathrm{M}$&
 			$\mathcal{D}_\mathrm{G} \rightarrow \mathcal{D}_\mathrm{D}$&
 			Avg\\
 			\hline
 			\hline
 			Global Li-Inter & 7.19& 	7.54&	7.29&	7.33&	7.34 \\
 			Bilinear-Inter& 6.78& 	7.31&	6.92&	6.9&	6.98  \\	
 			Ours & \textbf{6.45}& 	\textbf{7.22}&	\textbf{6.83}&	\textbf{6.86}&	\textbf{6.84} \\
 			\hline
 	\end{tabular}}
 	\caption{The performance of different interpolation methods.}
 	\label{tab:interpolation}
 \end{table}

 \begin{table}[h!]
 	\renewcommand\arraystretch{1.0}
 	\setlength{\tabcolsep}{2.2mm}
 	\centering
 	\scalebox{1}{
 		\begin{tabular}{|l|c|c|c|c|c|}
 			\hline
 			Number&
 			$\mathcal{D}_\mathrm{E} \rightarrow \mathcal{D}_\mathrm{M}$&
 			$\mathcal{D}_\mathrm{E} \rightarrow \mathcal{D}_\mathrm{D}$&
 			$\mathcal{D}_\mathrm{G} \rightarrow \mathcal{D}_\mathrm{M}$&
 			$\mathcal{D}_\mathrm{G} \rightarrow \mathcal{D}_\mathrm{D}$&
 			Avg\\
 			\hline
 			0&	7.00& 	7.44&	7.13&	7.26&	7.21 \\
 			64&	6.87& 	7.31&	7.08&	7.02&	7.07 \\
 			128&	6.66& 	7.21&	6.98&	7.05&	6.98 \\
 			256&	6.45& 	7.22&	6.83&	6.86&	6.84 \\
 			512&	\textbf{6.42}& 	7.19&	6.85&	6.81&	6.82 \\
 			1024&	\textbf{6.42}& 	\textbf{7.18}& 	\textbf{6.83}& 	\textbf{6.79}& 	\textbf{6.81} \\
 			\hline
 	\end{tabular}}
 	\caption{The Comparison for different number of negative samples.}
 	\label{tab:MCR_ablation}
 \end{table}

\subsubsection{Effectiveness of MCR Loss.}
To delve into the efficacy of our MCR loss, we setted varying numbers of negative samples to compare their cross-domain performance, as illustrated in Table \ref{tab:MCR_ablation}. When the number of negative samples is zero, it equates to a common contrastive regression loss, which is observed to achieve less than remarkable performance. As the number of negative samples increases from 64 to 256, there is a marked improvement in performance, indicating that a higher quantity of negative samples in contrastive learning contributes to performance gains. However, when the number of negative samples is further increased to 1024, the performance nearly reaches convergency, with minimal gains.

Moreover, we conducted an experiment on loss weights to explore the effectiveness of our method. For more details, please see the supplementary material.

\section{Conclusion}
In this paper, we propose a novel training framework named LG-Gaze for gaze estimation. Leveraging textual features extracted from language models, LG-Gaze guides the learning of gaze features. This approach enables the gaze model to acquire robust gaze representations and achieve a reasonable feature distribution, ultimately enhancing the cross-domain generalization capability of gaze model.
Our proposed LG-Gaze achieves state-of-the-art performance on
domain generalization for gaze estimation task.

\noindent\textbf{Acknowledgements.}
This work was sponsored by National Key R$\&$D Program of China (2023YFE0204200).

\bibliographystyle{splncs04}
\bibliography{main}

\end{document}